\documentclass[twoside]{article}
\usepackage{multirow}
\usepackage[para,online,flushleft]{threeparttable}
\usepackage{amsfonts}
\usepackage{amssymb}
\usepackage{amsmath}
\usepackage{mathtools}
\usepackage{caption}
\usepackage{booktabs}
\usepackage{array}
\usepackage{paralist}
\usepackage{bigstrut}
\usepackage{multirow}
\usepackage{dcolumn}
\usepackage{booktabs}
\usepackage{siunitx}
\usepackage{geometry}
\usepackage{graphicx}
\usepackage[utf8]{inputenc}
\usepackage[english]{babel}
\usepackage[numbers]{natbib}
\usepackage{multicol}
\providecommand{\U}[1]{\protect\rule{.1in}{.1in}}
\newtheorem{theorem}{Theorem}

\newtheorem{definition}[theorem]{Definition}

\newtheorem{lemma}[theorem]{Lemma}

\newenvironment{proof}[1][Proof]{\noindent\textbf{#1.} }{\ \rule{0.5em}{0.5em}}
\begin{document}

\title{Asymptotic Supervised Predictive Classifiers under Partition Exchangeability\thanks{I thank Prof. Jukka Corander for his approval of the results, Prof. Mark Daniel Ward for brainstorming on the paintbox process as Markov chain, and Dr. Jing Tang for providing the funding.}} 

\author{
  Ali Amiryousefi\footnote{Research Program in Systems Oncology, Faculty of Medicine, PO BOX 63, FI-00014,University of Helsinki, Finland. ali.amiryousefi@helsinki.fi}}
\date{}
\maketitle
\begin{abstract}

The convergence of simultaneous and marginal predictive classifiers under partition exchangeability in supervised classification is obtained. The result shows the asymptotic convergence of these classifiers under infinite amount of training or test data, such that after observing umpteen amount of data, the differences between these classifiers would be negligible.  This is an important result from the practical perspective as under the presence of sufficiently large amount of data, one can replace the simpler marginal classifier with computationally more expensive simultaneous one.

\textbf{Key words}: Supervised classification; Partition exchangeability; Sufficient statistic; Predictive classifier

\end{abstract}
\section{Background}
Under the broad realm of inductive inference, the goal of the supervised classification is to assign the test objects into \textit{a priori} defined number of classes learned from the training data  \cite{solo64}. One of the most applicable machinery that can optimally handle these scenarios is Bayesian which with a given prior information and accruing observed data, gradually enhances the precision of the inferred population's parameters \cite{hand01}. We consider here the general supervised classification case where the sets of species observed for features are not closed \textit{a priori}, leaving the probability of observing new species at any stage non-negative. The de Finetti type of exchangeability \cite{barlow92}, seems intractable in these cases. Nevertheless, one solution is to adhere to a form of partition exchangeability due to Sir John Kingman \cite{kingman78}. Assuming this type of exchangeability for each competing classes and harnessing the paintbox process as a Markov chain, the derivation here shows that given an infinite amount of data, the simultaneous and marginal predictive classifiers will be asymptotic. This is congruent with the similar study under the de Finetti exchangeability \cite{corander11}. Due to the existence of marginal dependency between the data points, the simultaneous and marginal classifiers are not necessarily equal. On the other hand, their convergence is not intuitive due to the complication posed with \textit{a priori} unfixed set of observable species. Upon availability of umpteen amount of data however, the proof presented here justifies the replacement of the marginal classifiers with the computationally expensive simultaneous ones. This is negating the divergence of these classifiers proved in the presence of only infinite training data \cite{corander13}. The following section introduces the predictive classifiers under partition exchangeability while the theorem denoting the convergence of these classifiers with recognition of paintbox process as Markov chain is presented in last section.

\section{Supervised predictive classifiers}

Consider the set of $m$ available training items by $M$ and correspondingly the
set of $n$ test items by $N$. For each item, we observe only one feature\footnote{One can assume that for each item we observe a finite vector of independent $d$ features, such that the element for the feature $j$ takes values in  $\mathcal{X}_{j},j=1,...,d$. WLOG, the derivation and results presented in this paper, considers $d=1$ and $\mathcal{X}_{1}=\mathcal{X}$.} that can take value from species set $\mathcal{X}=\{1,2, \ldots, r \}$. Note that each number in $\mathcal{X}$ is represented with one species such that the first species observed is represented with integer $1$, the second species is represented with integer $2$, and so on. On the other hand, $r$ is not known \textit{a priori} denoting the fact that we are uninformative about all of the species possible in our population.  
 A training item $i\in M$ is characterized by a feature $z_{i}$ such that, $z_{i}\in\mathcal{X}$. Similarly, we have for a
test item $i\in N$ the feature $x_{i}$ such that,
$x_{i}\in\mathcal{X}.$ Collections of the training and test data features
are denoted by vectors $\mathbf{z}$ and $\mathbf{x}$, respectively. Furthermore consider that the training data are allocated into $k$ distinct classes and $T$ is a joint labeling of all the training items into these classes. Simultaneous supervised classification will
assign labels to all the test data in $N$ in a joint manner. We can consider partitioning of $N$ test elements into $k$ different classes similar to $T$ such that $S=(s_{1},\ldots,s_{k}),s_{c}\subseteq N,c=1,...,k $ be the joint labeling of this partition. The $T$ and $S$ structures indicate a
partition of the training and test feature vectors, such that $\mathbf{z}^{(c)}$ and $\mathbf{x}^{(c)}$ represent the subset of training and test items in class $c=1,...,k$, respectively. The $\mathbb{S}$ denote the space of possible simultaneous classifications for a given $N$ and so $S \in \mathbb{S}$.

\subsection{Predictive classifiers}\label{PredClas}

Given the $M$ training data and their corresponding labels, the goal is to predict the classes that each  existing item in $N$ belongs to. A \textit{simultaneous} classifier of labels over $\mathbb{S}$ is defined as
\begin{equation}
p(S|\mathbf{x},\mathbf{z},T)=\frac{p(\mathbf{x}%
|\mathbf{z},S,T)p(S|T)}{\sum_{S\in\mathbb{S}}p(\mathbf{x}%
|\mathbf{z},S,T)p(S|T)},\label{posterior}
\end{equation}
where $p(S|T)$ is the uniform prior distribution and  
\begin{equation}
\mathcal{S}=p(\mathbf{x}|\mathbf{z},S,T),\label{S}
\end{equation} 
is the conditional predictive
probability for the whole observed population of test data. On the other hand, a
\textit{marginal} classifier specifies the predictive probabilities independently for
each test item, such that the conditional predictive distribution of the test data $p(\mathbf{x}|\mathbf{z},S,T)$ becomes
\begin{equation}
\mathcal{M}_{si}=\prod_{c=1}^{k}\prod
_{i: S_i\in c}p(x_{i}|\mathbf{z}^{(c)},T^{(c)}, S_i=c),\label{Msi}
\end{equation}
where $\mathbf{z}%
^{(c)}$ is the training data for class $c$. Note that this predictive distribution is constructed with the implicit assumption of availability of all the test data simultaneously. In the case that we are not provided with all the test data set but the test items are arriving \textit{sequentially} with the urge of being classified in each step, we need to construct $p(x_i|\mathbf{z}, T, S_i)=\prod_{c=1}^{k}p(x_{i}|\mathbf{z}^{(c)},T^{(c)}, S_i=c),$ 
for the $i$th test item. Hence, the predictive distribution of all the test data given their labels and training data $p(\mathbf{x}|\mathbf{z},S,T)$ in this case will be
\begin{equation}
\mathcal{M}_{se}=\prod
_{i=1}^{n}\prod_{c=1}^{k}p(x_{i}|\mathbf{z}^{(c)},T^{(c)}, S_i=c).\label{Mse}
\end{equation}

\subsection{Predictive classifiers under partition exchangeability}

Assume that number of species related to our feature is unfixed \textit{a priori}. Upon availability of the vector of test labels $S$, under \textit{partition exchangeability} framework, we can deduce the sufficient statistic for each subset of data. To define this statistic, consider the assignment of arbitrary permutation of integers $1,...,|s_{c}|$ to the items in $s_c$ where $n_{c}=|s_{c}|$ is the size of a given class $c$. Introducing the $I(\cdot)$ indicator function and $n_{cl}=\sum_{i\in s_{c}}I(x_{i}=l)$ as the frequency of items in class $c$ having value
$l \in \mathcal{X}$, then in terms of count in $\mathbf{x}^{(c)} $ one can write the sufficient statistic as
\begin{equation}\label{sufficient.stat.1}
\rho_{ct}=\sum_{l=1}^{\infty}I(n_{cl}=t),
\end{equation}
The vector of sufficient statistic $\mathbf{\rho
}_{c}=(\rho_{ct})_{t=1}^{n_{c}}$ indicates a partition of the integer $n_c$ such that $\rho_{ct}$ is the frequency of specific feature values that have been observed only $t$ times in class $c$ of test data. Given the above formulation, John Kingman, \cite{kingman77}, defined the random partition to be exchangeable if and only if when two different sequences having the same vector of sufficient statistics, have the same probabilities. Also in a representation theorem \cite{kingman78}, he showed that the probability distribution of the vector of sufficient statistic under this type of exchangeability will follow the Poisson-Dirichlet($\psi$) distribution known also as the Ewens sampling formula \cite{ewens72},
\begin{equation}\label{partitionexchangeability}
p(\mathbf{\rho})=\frac{n!}{\psi(\psi+1)\cdot\cdot\cdot(\psi+n%
-1)}\prod_{t=1}^{n}\left\{  (\frac{\psi}{t})^{\rho_{t}}\frac{1}%
{\rho_{t}!}\right\},  \quad \forall \quad \psi \in \mathbb{R}^+ , \quad \rho \in \mathfrak{S}_\rho,
\end{equation}
where,
\begin{equation}
\mathfrak{S}_\rho= \left\{(\rho_1, \rho_2, \ldots \rho_n) \quad | \quad \sum_{i=1}^n i\rho_i=n,\quad \rho_i \in \mathbb{N}_0, \quad i=1,2, \ldots n \right\}.
\end{equation}
Assuming the exchangeability of partitioning classes given $S$, the
product predictive for all test data is
\begin{equation}
p(\mathbf{x}|S)=\prod_{c=1}^{k}\frac{n_{c}!}{\psi
(\psi+1)\cdot\cdot\cdot(\psi+n_{c}-1)}\prod_{t=1}^{n_{c}}\left\{  (\frac{\psi
}{t})^{\rho_{ct}}\frac{1}{\rho_{ct}!}\right\}  .
\end{equation}
Analogously define $m_{c}$ and $m_{cl}$ as $n_{c}$ and $n_{cl}$, respectively, the vector of sufficient statistic for training and test data together $\tilde{\rho}_{c}=(\rho_{ct})_{t=1}^{n_{c}+m_{c}}$, can be represented by 
\begin{equation}\label{sufficient.stat2}
\tilde{\rho}_{ct}=\sum_{l=1}^{\infty}I(m_{cl}+n_{cl}=t).
\end{equation}
Before expressing the predictive classifiers with (\ref{partitionexchangeability}), define $\tilde{\rho}_{ct}{}^{(i)}$ as the updated sufficient statistic when data $x_{i}$ from only a
single test item is taken into account. More precisely, 
\begin{equation}\label{sufficient.stat3}
\tilde{\rho}_{ct}{}%
^{(i)}=\sum_{l=1}^{\infty}I(m_{cl}+n_{i;cl}=t),
\end{equation}
where $t=1,\ldots,m_{c}+1 $, and $n_{i;cl}$ is the observed frequency of category $l$ in item
$i$. 
Now with the use of tuned version of (\ref{sufficient.stat.1}) for the training data, (\ref{sufficient.stat2}), and (\ref{sufficient.stat3}), based on the Bayes' theorem, the updated version of predictive classifiers (\ref{S}) and (\ref{Msi}) is; 

\begin{gather}
\mathcal{S}=\frac{p(\mathbf{x}%
,\mathbf{z}|S,T)}{p(\mathbf{z}|T)}=\prod_{c=1}^{k}%
 \left\{\frac{
\frac{(m_{c}+n_{c})!}{\psi(\psi+1)\cdot\cdot\cdot(\psi
+m_{c}+n_{c}-1)}\prod_{t=1}^{m_{c}+n_{c}}\left\{  (\frac{\psi}{t}%
)^{\tilde{\rho}_{ct}}\frac{1}{\tilde{\rho}_{ct}!}\right\}} {\frac{m_{c}!}{\psi(\psi+1)\cdot\cdot\cdot(\psi+m_{c}%
-1)}\prod_{t=1}^{m_{c}}\left\{  (\frac{\psi}{t})^{\rho_{ct}}\frac{1}%
{\rho_{ct}!}\right\}  }\right\}.\label{S1}
\end{gather}

\begin{equation}
\begin{split}
\mathcal{M}_{si} & =\prod_{c=1}^{k}\prod
_{i: S_i\in c}\frac{p(x_i%
,\mathbf{z}^{(c)}|S_i=c,T^{(c)})}{p(\mathbf{z}^{(c)}|T^{(c)})}\\
& =\prod_{c=1}^{k}
\prod_{i: S_i\in c}%
 \left\{\frac{
\frac{(m_{c}+1)!}{\psi(\psi+1)\cdot\cdot\cdot(\psi
+m_{c}+1-1)}\prod_{t=1}^{m_{c}+1}\left\{  (\frac{\psi}{t}%
)^{\tilde{\rho}_{ct}^{(i)}}\frac{1}{\tilde{\rho}_{ct}^{(i)}!}\right\}} {\frac{m_{c}!}{\psi(\psi+1)\cdot\cdot\cdot(\psi+m_{c}%
-1)}\prod_{t=1}^{m_{c}}\left\{  (\frac{\psi}{t})^{\rho_{ct}}\frac{1}%
{\rho_{ct}!}\right\}  }\right\}.\label{Msi1}
\end{split}
\end{equation}
For writing the sequential marginal classifier $\mathcal{M}_{se}$,  note that for the $i$th test item, we are multiplying all the classes predictive distributions. This will be followed with the multiplication of these probabilities for all the $n$ elements. In doing so, the only class that needs to be updated for item $i$ is the one that the label of this item is predicted to belong to. To be precise, with introducing another indicator function $I_c^{(i)}$ that equals 1 if $s_i=c$, (\ref{Mse}) can be written as,
\begin{equation}
\begin{split}
\mathcal{M}_{se} & = \prod
_{i=1}^{n}\prod_{c=1}^{k}\frac{p(x_i%
,\mathbf{z}^{(c)}|S_i=c,T^{(c)})}{p(\mathbf{z}^{(c)}|T^{(c)})}\\
& =\prod
_{i=1}^{n}\prod_{c=1}^{k}%
 \left\{\frac{
\frac{(m_{c}+I_c^{(i)})!}{\psi(\psi+1)\cdot\cdot\cdot(\psi
+m_{c}+I_c^{(i)}-1)}\prod_{t=1}^{m_{c}+I_c^{(i)}}\left\{  (\frac{\psi}{t}%
)^{\tilde{\rho}_{ct}^{(i)}}\frac{1}{\tilde{\rho}_{ct}^{(i)}!}\right\}} {\frac{m_{c}!}{\psi(\psi+1)\cdot\cdot\cdot(\psi+m_{c}%
-1)}\prod_{t=1}^{m_{c}}\left\{  (\frac{\psi}{t})^{\rho_{ct}}\frac{1}%
{\rho_{ct}!}\right\}  }\right\}.\label{Mse1}
\end{split}
\end{equation}

\section{Asymptotic equivalence}

\subsection{Paintbox process as Markov chain}

Zabell in \cite{zabel92} discusses that as Bernoulli and multinomial trials are the building blocks of a general exchangeable sequence, so is the paintbox process for a general exchangeable random partition. Here we decompose this process and distinguish two fundamental components that jointly introduce a probability measure over $\left[0,1\right]$. The \textit{discrete} component comprise to those vectors $\textbf{p}=(p_1, p_2, \ldots)$ such that $p_1\geq p_2, \ldots \geq 0$ and $\sum_i p_i \leq 1.$  Vectors with such properties are called ordered defective probability vectors and we show the infinite simplex of such vectors with $\nabla$. For example a geometric sequences of the form $\left\{a_\nu=\theta^\nu\right\}, \nu \in \mathbb{N}$, for each $\theta\in \left[0, 1/2\right]$ is a proper candidate of this simplex. What is left from the infinite sum of the elements of the discrete component will constitute to \textit{continuous} component that is $p_0=1- \sum_i p_i$. Before formally define this process, note that  for $\textbf{p} \in \nabla$, Ewens sampling formula (\ref{partitionexchangeability}) quantifies probabilities of sufficient statistics governed with $\pi=(p_0, \textbf{p})$, exactly the same way $(p, 1-p)$ is for Bernoulli distribution. 

\begin{definition}[Paintbox process]
A process that in each stage can generate either a continuous value from interval $ \mathrm{r} \subseteq \mathbb{R}$ with probability $p_0$ or a discrete value from the set $ \mathfrak{n} \subseteq \mathbb{N}$ with corresponding probability $\normalfont{\textbf{p}}=(p_1, p_2, \ldots)$ for each element of $\mathfrak{n}$ is called paintbox process. Furthermore, this process is continuous or discrete if $\normalfont{\textbf{p}}=\pmb{0}$ or $\normalfont{p_0}=0$ , respectively.
\end{definition}
Note that this definition is denoting the paintbox process as a stochastic process with state space $\mathfrak{T}=\{\mathrm{r},\mathfrak{n}\}$. One can consider this as a first order stationary Markov chain with transition matrix $P=(P_{ij})$ where $P_{ij}$ is the transition function of the chain and is equal with $p_j$ for $ i,j \in \mathbb{N}_0$. In geometric sequence $\left\{a_\nu \right\}$ presented earlier, $\theta=0$ and $\theta=1/2$ on both extremes present continuous and discrete paintbox processes, respectively. In continuous case, the state space of the process will reduce to $\mathfrak{T}=\{\mathrm{r}\}$ while subset of natural numbers can be considered as the state space of the discrete paintbox process \textit{i.e.} $\mathfrak{T}=\{\mathfrak{n}\}$. Note that this truncation of state space is due to equality of corresponding elements of vector $\pi$ with zero. In general the zero elements in $\pi$ will result in the elimination of their corresponding states. In $\left\{a_\nu\right\}$, every other intermediate value of $\theta \in (0, 1/2)$ will result in a (\textit{mixed}) paintbox process. The following lemma describes the behavior of this process in a long run when this process is regarded as a stationary Markov chain.

\begin{lemma}[Positive recurrent paintbox process]\label{Lemma}
A paintbox process with state space $\mathfrak{T}$ and transition matrix $P$ is positive recurrent.
\end{lemma}
\begin{proof}
Verification of $\pi P=\pi$ denotes $\pi$ as the stationary distribution. On the other hand based on each nonzero probability assigned for visiting each state in next step given we are in a specific state we conclude that the chain is irreducible. These facts suggest that the chain is positive recurrent \cite{HP72}.   
\end{proof}

Note that since $\lim_{n \rightarrow \infty} P_{ij}^n=p_j$ for $i,j \in \mathbb{N}_0$, $\pi$ is also a steady state distribution of the chain, denoting the independence of the chain status in long run from its initial state. Furthermore, the result of the lemma above is underlying the fact that all the existing states will be visited infinitely many times. This has different implications regarding continuous and discrete parts of the state space $\mathfrak{T}$. First that the integers in the set $\mathfrak{n}$ will be visited infinitely many times. Second, the interval $\mathrm{r}$ will be also visited infinitely many times but in each visit, an outcome different from the previous visit to that state will be obtained. This is so since the probability of sampling any given point on a continuous interval is zero. In other words, we visit the interval $\mathrm{r}$ for infinitely many times but the outcomes of these visits are never the same. Gathering the infinite outcomes of the process, the sufficient statistic based on partition exchangeability will thus symbolically result in $\rho_{cont.}=(\infty, 0, \ldots)$, $\rho_{disc.}=(0, \ldots, \infty)$, or $\rho_{mix.}=(\infty, 0, \ldots, \infty)$ if the process is continuous, discrete, or mixed, respectively. 

\subsection{Convergence of simultaneous and marginal classifiers}

Now we can consider the behavior of the simultaneous and marginal classifiers under partition exchangeability assumption. In general, these classifiers are not necessarily equal, even if the test data were \textit{i.i.d.} from the same generative distribution as the training data. Here with increasing amount of training data however, the represented theorem shows that the simultaneous and marginal classifiers will coincide under the classification model
arising from the partition exchangeability. An intuitive implication of this result is that we are decoding the true probabilities of the underlying paintbox process as we accumulate more and more data points such that finally, the process of this learning expressed through the represented sufficient statistic will cease and new observations will not contribute to any significant amount in our learning.

\bigskip

\begin{theorem}[Asymptotic equivalence of predictive probabilities for
simultaneous and marginal classifiers]
Suppose that $m_c$ grows monotonically as $m$ does, then under partition exchangeable sampling process, 
\[
\lim _{m \to \infty} \frac{\mathcal{S}}{\mathcal{M}_{si}}=1=\lim_{m \to \infty} \frac{\mathcal{S}}{\mathcal{M}_{se}}. \]
\end{theorem}
\begin{proof}
Consider the left hand side of the equation (denoted by $l$ subscript) which is the division of the (\ref{S1}) and (\ref{Msi1}). Further consider writing the fraction in two pieces, namely coefficient $\mathtt{C}$ and sufficient parts $\mathtt{S}$. In the coefficient part observe,

\begin{gather}
\mathtt{C}_l=\left(\prod_{c=1}^{k}%
 \frac{
\frac{(m_{c}+n_{c})!}{\psi(\psi+1)\cdot\cdot\cdot(\psi
+m_{c}+n_{c}-1)}} {\frac{m_{c}!}{\psi(\psi+1)\cdot\cdot\cdot(\psi+m_{c}%
-1)}  }\right) \cdot
\left( \prod_{c=1}^{k}
\prod_{i: S_i\in c}%
 \frac{
\frac{(m_{c}+1)!}{\psi(\psi+1)\cdot\cdot\cdot(\psi
+m_{c}+1-1)}} {\frac{m_{c}!}{\psi(\psi+1)\cdot\cdot\cdot(\psi+m_{c}%
-1)}  } \right)^{-1}.
\end{gather}
Since the inner term of the right side is independent of $i$ we have,

\begin{gather}
\mathtt{C}_l=\left(\prod_{c=1}^{k}%
 \frac{
\frac{(m_{c}+n_{c})!}{\psi(\psi+1)\cdot\cdot\cdot(\psi
+m_{c}+n_{c}-1)}} {\frac{m_{c}!}{\psi(\psi+1)\cdot\cdot\cdot(\psi+m_{c}%
-1)}  }\right)\cdot
\left( \prod_{c=1}^{k} \left(
 \frac{
\frac{(m_{c}+1)!}{\psi(\psi+1)\cdot\cdot\cdot(\psi
+m_{c}+1-1)}} {\frac{m_{c}!}{\psi(\psi+1)\cdot\cdot\cdot(\psi+m_{c}%
-1)}  } \right)^{n_c} \right)^{-1}.
\end{gather}
$\mathtt{C}_l$ is the multiplication of all the classes coefficients as $\mathtt{C}_l=\mathtt{C}_{l_1} \cdot \mathtt{C}_{l_2} \cdots \mathtt{C}_{l_k}$. After simplifying the terms, for an arbitrary class $c$, the $\mathtt{C}_{l_c}$ can be written as,
\begin{gather}
\mathtt{C}_{l_c}=\frac{(\psi + m_c)^{n_c}}{(\psi + m_c)(\psi + m_c +1 ) \cdots (\psi + m_c +n_c -1)} \cdot \frac{(m_c!)^{n-1} (m_c+n_c)!}{(m_c +1)!^{n_c}}\\
=\frac{(\psi + m_c)^{n_c}}{(\psi + m_c)(\psi + m_c +1 ) \cdots (\psi + m_c +n_c -1)}\cdot \frac{(m_c+1)(m_c+2)\cdots (m_c+n_c)}{(m_c+1)^{n_c}}
.
\end{gather}
It is obvious now that tending $m_c$ to infinity, the two fractions comprising to $\mathtt{C}_{l_c}$ will be 1. Repeating this for the other classes leads to,
\[\lim_{m \to \infty} \mathtt{C}_l=1.\]
Before starting with the sufficient part, let's show the same result for the coefficient of the right hand side,
\begin{gather}
\mathtt{C}_r=\left(\prod_{c=1}^{k}%
 \frac{
\frac{(m_{c}+n_{c})!}{\psi(\psi+1)\cdot\cdot\cdot(\psi
+m_{c}+n_{c}-1)}} {\frac{m_{c}!}{\psi(\psi+1)\cdot\cdot\cdot(\psi+m_{c}%
-1)}  }\right) \cdot
\left( \prod_{i=1}^n
\prod_{c=1}^{k}%
 \frac{
\frac{(m_{c}+I_c^{(i)})!}{\psi(\psi+1)\cdot\cdot\cdot(\psi
+m_{c}+I_c^{(i)}-1)}} {\frac{m_{c}!}{\psi(\psi+1)\cdot\cdot\cdot(\psi+m_{c}%
-1)}  } \right)^{-1}.
\end{gather}
Like $\mathtt{C}_l$, this time with rearranging the terms in the right side of $\mathtt{C}_r$, verifies that $\mathtt{C}_r=\mathtt{C}_{r_1} \cdot \mathtt{C}_{r_2} \cdots \mathtt{C}_{r_k}$ and again each of the coefficients in each class will tend to 1 as $m_c$ tends to infinity, \textit{i.e.}
\[\lim_{m \to \infty} \mathtt{C}_r=1.\]
With the same approach, write the sufficient part of the sides as $\mathtt{S}_l =  \mathtt{S}_{l_1} \cdot \mathtt{S}_{l_2} \cdots \mathtt{S}_{l_k}$ and $\mathtt{S}_r =  \mathtt{S}_{r_1} \cdot \mathtt{S}_{r_2} \cdots \mathtt{S}_{r_k}$ which $\mathtt{S}_{._c}$ for an arbitrary class $c$ is equal with
\begin{gather}
\mathtt{S}_{l_c}= \frac{\mathtt{S}_{\mathcal{S}^c}}{\mathtt{S}_{\mathcal{M}_{si}^c}}
=\left(\frac{\prod_{t=1}^{m_{c}+n_{c}}\left\{  (\frac{\psi}{t}%
)^{\tilde{\rho}_{ct}}\frac{1}{\tilde{\rho}_{ct}!}\right\}}{\prod_{t=1}^{m_{c}}\left\{  (\frac{\psi}{t})^{\rho_{ct}}\frac{1}%
{\rho_{ct}!}\right\}}\right) \cdot \left(\prod_{i:S_i \in c} \frac{\prod_{t=1}^{m_{c}+1}\left\{  (\frac{\psi}{t}%
)^{\tilde{\rho}_{ct}^{(i)}}\frac{1}{\tilde{\rho}_{ct}^{(i)}!}\right\}}
{ \prod_{t=1}^{m_{c}}\left\{  (\frac{\psi}{t})^{\rho_{ct}}\frac{1}%
{\rho_{ct}!}\right\}}
\right)^{-1},\label{Sl}
\end{gather}

\begin{gather}
\mathtt{S}_{r_c}= \frac{\mathtt{S}_{\mathcal{S}^c}}{\mathtt{S}_{\mathcal{M}_{se}^c}}
=\left(\frac{\prod_{t=1}^{m_{c}+n_{c}}\left\{  (\frac{\psi}{t}%
)^{\tilde{\rho}_{ct}}\frac{1}{\tilde{\rho}_{ct}!}\right\}}{\prod_{t=1}^{m_{c}}\left\{  (\frac{\psi}{t})^{\rho_{ct}}\frac{1}%
{\rho_{ct}!}\right\}}\right) \cdot \left(\prod_{i=1}^n \frac{\prod_{t=1}^{m_{c}+I_c^{(i)}}\left\{  (\frac{\psi}{t}%
)^{\tilde{\rho}_{ct}^{(i)}}\frac{1}{\tilde{\rho}_{ct}^{(i)}!}\right\}}
{ \prod_{t=1}^{m_{c}}\left\{  (\frac{\psi}{t})^{\rho_{ct}}\frac{1}%
{\rho_{ct}!}\right\}}
\right)^{-1}.
\end{gather}
It is obvious that the second parentheses in $\mathtt{S}_{l_c}$ is equal to its corresponding value in $\mathtt{S}_{r_c}$ since for the items that are not in a specific class $c$, $I_c^{(.)}$ will be zero and $\tilde{\rho}_{ct}^{(i)}$ will be the same as $\rho_{ct}$. This means $\mathtt{S}_{\mathcal{M}_{si}^c}=\mathtt{S}_{\mathcal{M}_{se}^c}$. So essentially we have the equality of the sufficient parts in each class that consequently means $\mathtt{S}_l=\mathtt{S}_r$.
Hence by showing that (\ref{Sl}) tends to 1 if $m$ and consequently $m_c$ tends to infinity, the proof is complete and this is so since in this case, the three sufficient statistics $\tilde{\rho}_{ct}, \tilde{\rho}_{ct}^{(i)}$ and ${\rho}_{ct}$ will converge based on the result of the presented lemma. 
\end{proof}

Note that equally tending $n$ and consequently $n_c$ to infinity with the same line of logic still leads to the equivalent result as above. 
 

{\noindent}
\end{document}